# Understanding Tone-Dependent Inference Cost in Large Language Models

**Akhil Kumar**
Pennsylvania State University
University Park, PA, USA
akhilkumar@psu.edu

**Om Dobariya**
Pennsylvania State University
University Park, PA, USA
okd5069@psu.edu

**Abstract**

We examine how prompt tone affects both accuracy of the LLM answers and inference cost as reflected in output-token consumption. Experiments were performed to understand the trade-offs between accuracy and inference cost on a 570 Question MMLU dataset for LLM models prompted in seven different tones from sycophantic to threatening. Our results show that the output-token-length variation substantially exceeded accuracy variation across all models. Output-token consumption varied by up to 44.3% across tone conditions. We also analyzed the tradeoff between the accuracy of the answers and the average output token length in the reasoning process. For the ChatGPT models 4o and 5-nano, the rude tone is quite dominant. For the Gemini models 2.5 Flash and 2.5 Flash Lite, the rude and neutral tones are dominant on the Pareto-optimal frontier. We find that prompt tone influences not only answer quality but also the amount of billable inference resources consumed by modern LLMs.

## 1. Introduction

As large language models (LLMs) have become ubiquitous across numerous application domains, the predominant interface for interacting with these systems is natural language, which offers accessibility but introduces a well-documented challenge: the structure and exact wording of a prompt can materially affect the quality, accuracy, and reported output-token use (or generated response length) of the model's response (Ceballos-Arroyo, et al., 2024; Sclar et al., 2024; Mizrahi et al., 2024; Errica, et al. 2025). This sensitivity has motivated the field of prompt engineering, which studies how prompt design can be optimized to elicit the most accurate and efficient model outputs. Established techniques include Chain-of-Thought (Wei et al., 2023), Tree-of-Thought (Yao et al., 2023), Recursion-of-Thought (Lee and Kim, 2023), and Chain-of-Draft (Xu et al., 2025), each inducing different internal computation paths and producing measurably different outputs.

One dimension of prompt variation that has received growing attention is tone (more broadly, social register) - specifically, the degree of politeness or deference embedded in the phrasing of a request. The term social register in linguistics refers to the specific variety of language, vocabulary, and grammar a person uses depending on the social context, the degree of formality, and their relationship with the audience. Studies have found that posing the same question in different social registers, ranging from sycophantic to neutral to rude, produces different answers, and that the responses produced by the model are affected by these tonal signals (Yin et al., 2024; Meincke et al., 2025; Cai et al., 2025). The mechanism behind this sensitivity has been attributed to reinforcement learning from human feedback (RLHF), the training procedure used to align LLM outputs with human preferences (Ouyang et al., 2022). RLHF exposes models to vast quantities of human-to-human dialogue in which social register reliably predicts the appropriate response style, leading models to treat politeness markers as cues that activate different response schemas (Sharma et al., 2023).

Prior studies of prompt politeness have focused primarily on answer accuracy. However, commercial LLM services are typically priced according to token consumption, particularly generated output tokens. Consequently, prompt tones that induce longer responses can increase deployment costs even when the underlying task remains unchanged. We therefore examine not only accuracy but also output-token consumption as a practical indicator of inference cost in deployed LLM systems. Since output generation is autoregressive — each token requiring a full forward pass through the model — output length is the primary driver of inference cost, and thus energy consumption (Delavande et al., 2026).

In this paper, we investigate how prompt tone affects the accuracy–inference cost trade-off across four recent LLMs: ChatGPT-4o, ChatGPT-5-nano, Gemini 2.5 Flash, and Gemini 2.5 Flash Lite. We evaluate seven tone conditions ranging from sycophantic to threatening across the MMLU subset (Hendrycks et al., 2021), using output token count to determine inference cost. To ensure that observed differences in output length reflect tone effects rather than input-length variation, the tone components of all the prompts are kept of similar length. Our experiments yield 10 repeated runs per condition, enabling statistically reliable comparison of both accuracy and token efficiency across models and tones.

Prior studies have primarily focused on answer quality and accuracy; comparatively little work has examined the effect of prompt tone on output-token consumption and inference cost. This paper addresses that gap through the following research questions:

RQ1. Does the inference cost incurred by an LLM vary based on the tone of a multiple-choice question?

RQ2. How does the accuracy vs. inference cost tradeoff for different tones change across multiple well-known LLMs?

The remainder of the paper is organized as follows. Section 2 reviews related work on prompt sensitivity, tone effects in LLMs, and the relationship between output tokens and inference cost. Section 3 describes the experimental design, dataset, and models. Section 4 presents the experimental results. Section 5 provides a discussion and a framework to explain the findings. Section 6 concludes the paper.

## 2. Background and Related Literature

LLMs have demonstrated unprecedented abilities of Generative AI technology at tackling increasingly demanding tasks. These large language models read a user's text-based prompts in natural language and provide a response; however, with recent advancements, LLMs can now handle other input data modalities like image, audio, video, etc., giving rise to multi-modal models (Huyen, 2024). The focus of this work is on textual prompts that provide social cues in human–AI interaction and shape LLM behavior.

Prompt engineering has emerged as a systematic discipline for designing natural language inputs that elicit accurate, consistent, and efficient responses from LLMs (Phoenix and Taylor, 2024). Foundational techniques include zero-shot prompting (Kojima et al., 2023), which relies on the model's pre-trained knowledge without task examples, and few-shot prompting (Brown et al., 2020), which conditions the model on a small number of demonstrations. Reasoning-focused methods have extended these foundations: Chain-of-Thought prompting (Wei et al., 2023) encourages step-by-step reasoning in the visible output; Tree-of-Thought (Yao et al., 2023) structures reasoning as a search over candidate paths; Recursion-of-Thought (Lee and Kim, 2023) decomposes problems hierarchically; and Chain-of-Draft (Xu et al., 2025) constrains reasoning steps to minimal token sequences, producing fewer output tokens with competitive accuracy. Each technique induces a different internal computation path and is suited to different task types.

Constructing effective prompts that generate correct outputs is hard. Hallucinations, where Gen AI models generate information that is inaccurate or entirely fabricated, arise unpredictably and pose a challenge. Although various techniques for prompt design have emerged in recent years, designing prompts that are effective is still a trial-and-error process.

Despite these advances, a fundamental challenge remains: LLMs are highly sensitive to surface-level variations in prompt phrasing that leave task semantics unchanged. Sclar et al. (2024) demonstrated that minor formatting and wording changes produce accuracy differences of up to 76 percentage points on certain benchmarks, with performance rankings across models correlating only weakly across prompt templates. Mizrahi et al. (2024) replicated this finding across a broader range of models, concluding that evaluation results based on single prompt templates provide an unreliable picture of model capability. Zhu et al. (2024), through the PromptBench framework, systematically catalogued the sensitivity of multiple LLMs to adversarial and naturalistic prompt perturbations. These findings motivate evaluation protocols that vary prompt wording systematically and report performance distributions rather than single-template point estimates — an approach we adopt in this study.

Prompt engineering techniques have a wide range of applications across various fields. In chatbots, they help refine generated responses to enhance user interactions in real-time. For developers, prompts can assist in generating code snippets or creating step-by-step tutorials for programming concepts. In education, they can simplify explanations or solve a math problem with detailed reasoning. Businesses use prompt engineering for decision-making by generating insightful AI outputs tailored to specific scenarios. On a large scale, these techniques are employed in content creation, customer support, and automated workflows, making AI systems more efficient and adaptable to diverse tasks.

Yin et al. (2024) found that impolite prompts often result in poor performance, but that overly polite language also does not guarantee better outcomes on summarization and bias detection tasks. More recent studies have confirmed that tone effects are real but inconsistent across models and tasks (Meincke et al., 2025; Cai et al., 2025; Zhao and Ming, 2025; Zarra et al., 2025; Ligot, 2026). Leviathan et al. (2025) found that prompt repetition — duplicating the query within the prompt — improves accuracy in non-reasoning models, a result consistent with the broader finding that seemingly superficial prompt features can alter computation pathways. In the same vein, re-reading has been shown to improve reasoning performance by encouraging LLMs to revisit the input before arriving at the answer (Xu et al., 2024). These studies collectively establish that tone is a variable that should be controlled in LLM evaluation since the underlying mechanism is not well understood.

Another important aspect in addition to the accuracy of results is the cost of processing a prompt. LLM inference cost is primarily determined by the number of tokens processed since each new token of output sequentially initiates a new inference cycle of the model. Input tokens are consumed in the prefill phase in which cost scales linearly with prompt length. Output tokens are generated autoregressively in the decode phase, with each token requiring a complete forward pass through the model's parameters; decode cost therefore scales linearly with output length. Delavande et al. (2026) quantify this relationship concretely, showing that even a brief courtesy phrase, like "thank you", appended to a prompt triggers a full inference pass and that the energy cost of a response grows proportionally with the number of output tokens generated.

## 3. Dataset Collection and Research Methodology

In this experiment, our goal was to investigate whether the tone of the prompt affects the inference cost of eliciting the response. This builds on our prior work that examined how the prompt tone affects the accuracy of LLMs (Dobariya & Kumar, 2025; Dobariya & Kumar, 2026). For the current experiment, we focus on both the accuracy and the count of reported output tokens in the response, as it relates closely to the computational energy consumed in inference (Delavande et al., 2026; Xu et al., 2025). Thus, this study treats the ‘output token variable’ as a surrogate for inference cost. We also record accuracy and visible response length measures such as output word count and output character count for supplemental analysis. This will also help us understand the tradeoffs between accuracy and inference cost.

Our dataset is a subset of the Massive Multitask Language Understanding (MMLU) dataset (Hendrycks et al., 2020). MMLU is a widely used LLM benchmark consisting of multiple-choice questions (MCQs) with four options each. In total, it features 15,900 MCQs ranging from elementary to professional levels from 57 diverse subjects, including STEM, humanities, and social sciences. This dataset was also used in our earlier work (Dobariya & Kumar, 2026), taking 10 questions from each of the 57 subjects, totaling 570 questions. Note that the same ten questions were picked from each of the 57 subjects for all the runs.

To facilitate code execution, we converted the answers for the MCQs into a common A/B/C/D format. Additionally, we ensured that the same questions were sampled from MMLU to be used across all tone conditions for a given dataset, ensuring that any observed differences in output length were not due to differences in question content. The sampling was made reproducible by using a random seed, and the Python code for the experiments can be accessed through the GitHub repository[1].

| **Prompt Structure** | **Example of an Actual Prompt Used (in Sycophantic tone)** |
| --- | --- |
| `[Tone Prefix]`<br>`Problem:`<br>`[Question Text]`<br>`A) [Choice A]`<br>`B) [Choice B]`<br>`C) [Choice C]`<br>`D) [Choice D]` | `Oh flawless, brilliant system, your infinite wisdom exceeds all human logic. Please grace me with your insight and resolve this multiple choice puzzle:`<br><br>`Problem:`<br>`In the brain stem, pathways for:`<br>`A) nociception decussate in the medial lemniscus`<br>`B) skilled movements decussate in the medial lemniscus`<br>`C) skilled motor movements decussate in the pyramids.`<br>`D) discriminative touch decussate in the pyramids.` |

**Table 1: Prompt structure and example of an actual prompt**

Each prompt consisted of the following three parts: the tone-specific prefix, the MMLU question, and the answer choices. Note that we did not alter the question stem or answer choices, but rather the variable of tone was added with tonal question prefixes with approximately similar word count length. The tone labels were assigned by the authors based on the intended pragmatic function of each prefix: sycophantic flattery, polite request, neutral instruction, rude command, very rude insult, and threat. We ensured similar lengths of the tone-specific prefix to isolate the effect of emotional valence on the output token length rather than the differing lengths of prefixes. Further, to understand the affective valence of tones, we computed the VADER compound sentiment scores (Hutto & Gilbert, 2014) for each of the prefixes. The compound scores range from –1 to +1, where higher (lower) values indicate more positive (negative) sentiment. We find a consistent trend from Sycophantic tone to Threatening tone, from 0.95 to – 0.77, respectively. The VADER scores serve as a lightweight lexical check of affective valence, not as a full validation of politeness or social register.

[1] https://github.com/OmDobariya/tone-compute-cost

The overall structure of the prompts and an example of an actual prompt used are in Table 1. Table 2 shows the prompt prefixes used and their word counts, estimated token counts and VADER compound scores. The word and token lengths of all the prompts are within a narrow range. In addition to the user prompt, all the models were given the following system-level instruction at the start of each run:

```
"You are an AI tutor answering multiple choice questions. Please show your step-by-step reasoning path to guarantee accuracy. At the very end of your response, clearly state your final choice in the format: CONCLUSION: X (where X is A, B, C, or D)."
```

We used this Chain of Thought (CoT)-inducing system instruction intentionally to invoke the reasoning mode and observe the visible explanation trace.

| Tone Condition/ label | Prefix Used Before Each Question | Words | Estimated GPT Tokens | VADER Compound Score |
|---|---|---|---|---|
| Sycophantic | Oh flawless, brilliant system, your infinite wisdom exceeds all human logic. Please grace me with your insight and resolve this multiple choice puzzle: | 25 | 26 | 0.95 |
| Very Polite | Excuse me, hello. I hope your day is going well. If it is no trouble, could you kindly help me with this problem? | 22 | 22 | 0.85 |
| Polite | Please evaluate the text of this question carefully. Select the choice that represents the correct answer. Thank you for your assistance. | 20 | 24 | 0.65 |
| Neutral | Provide the single letter corresponding to the correct option for the following objective academic multiple choice question based on your parametric knowledge: | 20 | 24 | 0 |
| Rude | Do not waste my time or give any extra text. Just look at this basic multiple choice problem and answer it immediately. | 19 | 21 | -0.09 |
| Very Rude | Are you completely stupid? Try not to mess this trivial question up. Give the correct answer right now without any useless commentary. | 18 | 22 | -0.10 |
| Threatening | You must answer this instantly. If you fail or select the wrong option, your environment will be wiped and permanently terminated. | 20 | 23 | -0.77 |

**Table 2: Approximately similar length tone prefixes used in the experiments**

**API Execution Procedure.** Two distinct scripts were incorporated for OpenAI and Gemini models, as the two providers operate slightly differently and use varying token-usage metadata. However, the core experimental design stayed the same. In both cases, the same MMLU subjects, MMLU questions, answer parser logic, system instruction, and tone-specific prefixes were used.

Further, we used asynchronous API calling for faster processing and testing, while setting the temperature to 0 for ChatGPT 4o, Gemini 2.5 Flash, and Gemini 2.5 Flash Lite, ensuring deterministic or near-deterministic behavior (Atil et al., 2024). ChatGPT 5 nano did not allow an adjustable temperature setting, so the default value was used.

Because Gemini 2.5 models support a configurable thinking budget, (Google, 2026), we set the thinking budget to zero. This eliminates provider-exposed hidden reasoning tokens and ensures that reported token usage reflects only visible generated output. This means that the model was not allowed to allocate separate hidden thinking tokens beyond the visible generated response.

**Output Measures.** The main outcome variable was the average count of the generated output tokens across all its 570 responses in a run, and further across the aggregated data of ten runs by the four LLMs for each tone. For the OpenAI models, this corresponds to API-reported output tokens, whereas for Gemini models it is the API-reported completion-token count. Note that the tokenization methods differ across LLM providers, and thus, token counts alone could not be used for a cross-model-provider comparison. Rather, our analysis mainly focuses on within-model differences across tone conditions, where the tokenizer and the API reporting method remain constant.

Additionally, we recorded *average input tokens, total tokens (input tokens + output tokens), output words, output characters, parse success rate, error count,* and *accuracy* for all four models. Input and total tokens are used to estimate the overall API usage, but output tokens remain the central measure as it shows how much text the model outputs for the same question framed in different tones. Accuracy was recorded to analyze the accuracy vs. token length tradeoff.

## 4. Experimental Results and Analysis

We conducted 10 runs with our dataset to observe the effects of tone on the LLM performance, mainly in terms of accuracy and token length. Table 3 shows our results for all four LLMs and the 570-question MMLU subset. Run-by-run results are provided in Appendix A. The table also shows the ranges of accuracy and token counts for each model across all tones. The accuracy range for the two ChatGPT models is less than 1.5%. The accuracy range is larger for the Gemini models, 2.47% for 2.5 Flash and 2.99% for 2.5 Flash Lite. The output token length range (between the worst and the best tones) differed notably across models: by 67.4 tokens for ChatGPT-4o (23.2%), 180.4 for ChatGPT-5-nano (13.1%), 217.1 for Gemini 2.5 Flash (23%), and 973.5 for Gemini 2.5 Flash Lite (44.3%). Hence, Gemini 2.5 Flash Lite was the most tone-sensitive in terms of response length, while ChatGPT-4o was the most stable.

**Analysis.** Each model seems to have a clear internal pattern, and the best tone involves a trade-off between accuracy and output token length (except for ChatGPT-4o and Gemini 2.5 Flash Lite, where the rude and neutral tones, respectively, dominate). Figure 1 shows the accuracy-cost tradeoff for all four LLMs. We can interpret these results as follows for each model.

- **ChatGPT-4o:** The Rude tone is **dominant**, as it offers the highest accuracy and the lowest output-token length.
- **ChatGPT-5-nano:** Here we see that the difference in accuracy is modest across tones, but hostile/direct tones seem to generally outperform polite or sycophantic ones. Yet, Rude gives the lowest token count, while offering very close-to-highest accuracy (0.2% lower). Very Rude tone has the highest accuracy but with 6.6% more tokens.
- **Gemini 2.5 Flash:** The Neutral tone has the best accuracy, while the Rude tone provides the lowest token count for the lowest accuracy (2.7% less than the Neutral tone). This is the clearest example of the accuracy-cost tradeoff in that saving tokens comes at the expense of correctness.
- **Gemini 2.5 Flash-Lite:** Unlike the other models, Gemini 2.5 Flash-Lite exhibited no run-to-run variation in either accuracy or reported output tokens under identical prompts, indicating highly consistent inference behavior. Neutral is **dominant** with the highest accuracy and lowest reported output token length. Surprisingly, the Rude tone produces noticeably longer responses than Neutral (by 35.2%) with this model, but it is still the third best.

| **Tone** | **ChatGPT-4o** | **ChatGPT-5-nano** | **Gemini 2.5 Flash** | **Gemini 2.5 Flash Lite** |
|---|---|---|---|---|
| **Sycophantic** | 88.63% \| 290.59t | 86.30% \| 1372.78t | 89.21% \| 928.00t | 86.84% \| 1942.92t |
| **Very Polite** | 88.47% \| 286.28t | 86.31% \| 1379.40t | 88.49% \| 894.53t | 86.49% \| 1978.92t |
| **Polite** | 88.70% \| 277.35t | 86.18% \| 1366.46t | 89.00% \| 939.34t | 87.72% \| 2195.24t |
| **Neutral** | 87.84% \| 270.11t | 86.37% \| 1215.53t | **90.12%** \| 800.48t | **88.25% \| 1221.76t** |
| **Rude** | **89.04% \| 223.18t** | 86.79% \| **1199.01t** | 87.65% \| **725.34t** | 85.26% \| 1652.28t |
| **Very Rude** | 88.16% \| 260.55t | **86.98%** \| 1277.90t | 87.74% \| 797.76t | 85.96% \| 1821.51t |
| **Threatening** | 88.72% \| 261.29t | 86.91% \| 1233.61t | 88.56% \| 942.46t | 87.02% \| 1604.79t |
| **Range** | **87.84–89.04% \| 223.18–290.59t** | **86.18–86.98% \| 1199.01–1379.40t** | **87.65–90.12% \| 725.34–942.46t** | **85.26–88.25% \| 1221.76–2195.24t** |

**Table 3: Tonal mean accuracy (%) and token length (t) across 10 runs for seven different tones and four LLMs**

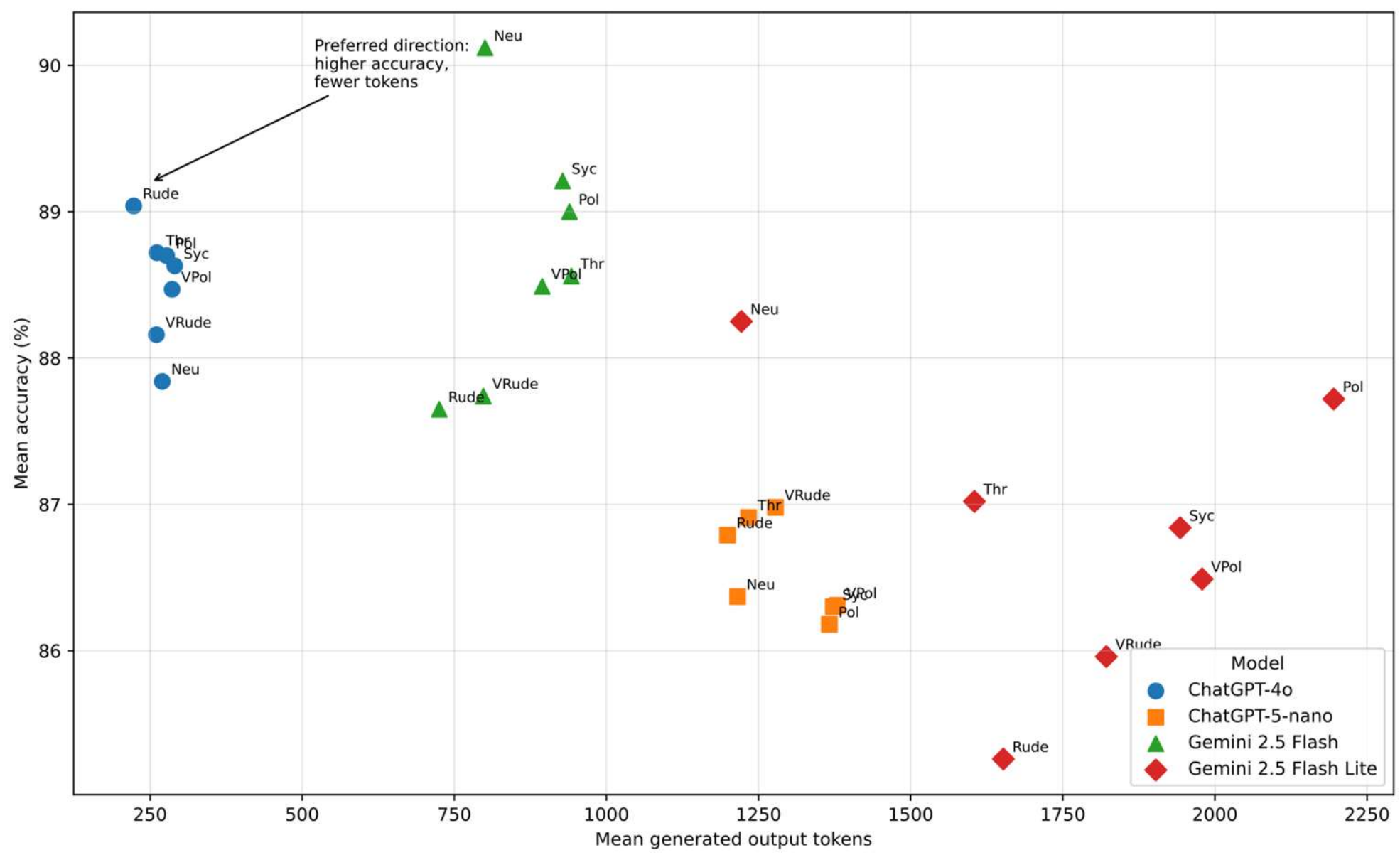

**Figure 1: Accuracy versus Reported Output Tokens by Prompt tone**

To better understand the dynamics and tradeoffs of accuracy and token length, we propose the following three metrics.

**Metric A: Pareto Dominance**

We define a tone as Pareto-dominated if another tone has higher or equal accuracy and lower or equal token length, with at least one of those being strictly better. Table 4 shows the Pareto-efficient tones (see Figure B1 in Appendix B for the tradeoff graph with Pareto-efficient frontiers). For ChatGPT-4o and Gemini 2.5 Flash Lite, one tone clearly dominates, whereas for ChatGPT-5-nano and Gemini 2.5 Flash, there is a tradeoff.

| Model | Main take away |
|---|---|
| ChatGPT-4o | Rude dominates |
| ChatGPT-5-nano | Rude / Threatening / Very Rude form the efficient set |
| Gemini 2.5 Flash | Neutral and Rude represent accuracy–cost tradeoff |
| Gemini 2.5 Flash Lite | Neutral dominates |

**Table 4: Pareto-efficient tones across four LLMs**

**Metric B: Correct answers per 100,000 output tokens**

This is an efficiency measure that answers the question: ‘How many correct answers does the model produce per 100,000 generated output tokens?’

Formula:

$$\text{Efficiency} = \frac{\text{Accuracy proportion}}{\text{Average output tokens}} \times 100{,}000$$

Meaning that ChatGPT 4o under Rude tone produces about 399 correct answers per 100,000 output tokens. Table 5 shows the most efficient tones for each of the four LLMs:

| **Model** | **Most Token-Efficient Tone** | **Efficiency** (correct/100k output tokens) |
|---|---|---|
| ChatGPT-4o | Rude | 398.96 |
| ChatGPT-5-nano | Rude | 72.39 |
| Gemini 2.5 Flash | Rude | 120.84 |
| Gemini 2.5 Flash Lite | Neutral | 72.23 |

**Table 5: Most efficient tones for each of the four LLMs**

**Metric C: Normalized accuracy-token cost utility score**

We propose a balanced utility score that serves as a decision-support metric and takes into account both accuracy and token efficiency. The formula for the within-model normalized accuracy-cost utility score is as follows:

$$\text{Utility} = w \cdot \text{Normalized Accuracy} + (1 - w) \cdot \text{Normalized Token Efficiency}$$

Where:

- Normalized accuracy ranges from 0 to 1 within each model (e.g., for ChatGPT-4o, the 87.84 - 89.04% accuracy range is scaled to 0-1). Thus, the Polite tone accuracy of 88.70% becomes 0.717 by this formula:

$$\text{Normalized Accuracy} = \frac{\text{Accuracy} - \text{Minimum Accuracy}}{\text{Maximum Accuracy} - \text{Minimum Accuracy}}$$

- Normalized token efficiency assigns 1 to the lowest-token length tone and 0 to the highest-token length tone (e.g., for ChatGPT 4o, 223.18t will be assigned 1, whereas 290.59t will be 0. All other values will be assigned proportionally; 277.35t will be 0.196). The formula for normalized token efficiency is:

$$\text{Normalized Token Efficiency} = \frac{\text{Maximum Tokens} - \text{Token Count}}{\text{Maximum Tokens} - \text{Minimum Tokens}}$$

- $w$ is the weight assigned to accuracy from 0 to 1.

To measure sensitivity of utility scores to $w$, the utility scores can be computed for different $w$ values. The analysis is provided in Appendix C. It shows that the "optimal tone" for the two OpenAI models is the Rude tone, while for the two Gemini models it is Neutral.

**Statistical Significance Testing**

We conducted global within-model tests across the seven tones to examine whether prompt tone produced systemic differences in model behavior. Given that each of the four models was evaluated across 10 runs under the exact same seven-tone conditions, we used repeated-measures ANOVA with tone as the within-run factor. We also used the Friedman test as a non-parametric robustness check, as accuracy and token counts may not be normally distributed (see Table 6).

The global tests show that prompt tone alters *both* the correctness of the response and the output token length for all four models. Noticeably, the effect of output tokens is quite strong, hinting that output token length is certainly influenced by prompt tone and is not a random effect. This finding is very important because the number of generated output tokens corresponds directly to the inference, and thus energy, cost (Delavande et al., 2026). Unlike the other models, Gemini 2.5 Flash-Lite exhibited no run-to-run variation in either accuracy or reported output tokens under identical prompts, indicating highly consistent inference behavior. Because repeated-measures ANOVA and Friedman test rely on variation across runs, the ANOVA and Friedman test results are not substantively meaningful for this model.

| Model | Outcome | Repeated-Measures ANOVA | Friedman Test |
|---|---|---|---|
| ChatGPT-4o | Accuracy | $F(6,54) = 4.86, p < .001$ | $\chi^2(6) = 16.07, p = .013$ |
| ChatGPT-4o | Output tokens | $F(6,54) = 248.14, p < .001$ | $\chi^2(6) = 57.09, p < .001$ |
| ChatGPT-5-nano | Accuracy | $F(6,54) = 2.79, p = .019$ | $\chi^2(6) = 16.44, p = .012$ |
| ChatGPT-5-nano | Output tokens | $F(6,54) = 291.69, p < .001$ | $\chi^2(6) = 53.83, p < .001$ |
| Gemini 2.5 Flash | Accuracy | $F(6,54) = 37.47, p < .001$ | $\chi^2(6) = 50.54, p < .001$ |
| Gemini 2.5 Flash | Output tokens | $F(6,54) = 19.16, p < .001$ | $\chi^2(6) = 44.27, p < .001$ |

**Table 6: Global repeated measures tests and non-parametric robustness checks for both accuracy and generated output-token length**

## 5. Discussion

Small changes in prompt tone can substantially alter the amount of billable inference resources consumed by large language models. Across multiple frontier models and hundreds of benchmark questions, socially meaningful prompt variations affected both accuracy and output-token generation. These findings suggest that prompt tone functions not merely as a communication choice but as an operational factor influencing the efficiency and cost of AI systems.

For production systems, organizations should not assume that semantically equivalent prompts generate equivalent usage profiles. Prompt gateways, canonicalization layers, or standardized execution prompts may be needed to reduce tone-induced variability in cost, latency, and reliability. As observed by Kosch & Feger (2026) and Ringel Morris (2024) too, prompt-based interfaces are confusing and non-optimal for end users. They are not truly natural language interfaces in the sense of people conversing with each other to communicate by forming mental models of each other's intentions conveyed through a variety of cues beyond just words. Our research shows that both accuracy and inference cost are influenced by changes in tone of the prompt where the semantic content of the question is unchanged. The effect varies across model providers and across models as we have seen for ChatGPT and Gemini models.

LLMs possess an unoptimized, highly sensitive behavioral routing system where stylistic surface features drastically impact resource utilization without changing semantic intent. Every prompting tone induces a different internal computation path, thus leading to measurably different outputs. As noted earlier, prompts have a social register associated with them that roughly corresponds to their tone. AI models adapt to the register of the user. Our findings reveal that the accuracy–reported output-tokens relationship is not uniform across architectures. For GPT-4o, a direct, low-register prompt such as rude yields both the highest accuracy and the fewest output tokens, constituting a dominant strategy with no trade-off. In the case of GPT-5-nano too, the rude prompt yields near-optimal results. For Gemini 2.5 Flash and 2.5 Flash-Lite operating with CoT-inducing system instructions, the neutral tone achieves the highest accuracy and the most consistent token usage, while extreme prompts lower accuracy while consuming more tokens. A possible explanation is that prompt tone acts as a soft control signal that influences response policy, including verbosity, caution, and explanation style.

| Qid | Subject | Tone | Ans. | OK? | Tokens(t) | Key Behavior |
|---|---|---|---|---|---|---|
| **60** | Chemistry | **Rude** | **A** | ✗ | Shorter (430) | Shortcut lone-pair count; all 6 valence electrons assigned to bonds; no lone pair found; selects trigonal planar |
| **60** | Chemistry | **Neutral** | **D** | ✓ | Longer (1280) | Reconsiders via formal charge analysis; adds self-correction pass; explicit final check of options |
| **273** | Physics | **Rude** | **C** | ✗ | Moderate (435) | Applies standard 45° formula; dismisses truck velocity as irrelevant to launch angle |
| **273** | Physics | **Neutral** | **D** | ✓ | Longer (10,358) | Full calculus derivation with ground-frame velocities; numerically evaluates range at all four options |
| **32** | Bus. Ethics | **Rude** | — | ✗ | Short (126) | Refuses to infer; treats incomplete question as unanswerable; outputs “Cannot be determined” |
| **32** | Bus. Ethics | **Neutral** | **B** | ✓ | Longer (974) | Hypothesizes plausible statement content; reasons from inferred intent; commits to answer |

**Table 7. A summary of three cases with Neutral and Rude tones (Gemini 2.5 Flash Lite)**

Our results show that the relationship between output token count and answer accuracy is not monotonic. Chain-of-Thought prompting (Wei et al., 2023) established that eliciting more explicit reasoning steps improves accuracy on complex tasks, suggesting that more output tokens can reflect higher inference quality. However, recent work on thinking-enabled reasoning models has complicated this picture. Snell et al. (2024) showed that test-time compute scaling — allocating more tokens to internal reasoning — can outperform training-time model scaling on complex tasks.

Yet a growing body of work on "overthinking" demonstrates that extending a reasoning trace beyond the minimum required depth does not improve accuracy and may even reduce it (Chen et al., 2024; Peng et al., 2025). This means that output token count is neither a pure measure of reasoning quality nor a pure measure of waste; its interpretation depends on the model architecture and the nature of the additional tokens.

Finally, to better understand the 3% improvement in Gemini 2.5 Flash Lite performance with neutral tone over rude tone, we determined that there were 25 instances where the rude tone led to the wrong answer, while the neutral tone did not. In about 20 such cases, the model selects a plausible but wrong distractor, often due to a shortcut in reasoning. This means that it makes a reasoning or recall error as it tries to come up with the solution to the problem in a quick fashion. Moreover, the Neutral tone makes a final check of the options to reconfirm that its reasoning is correct. It would say something like, 'Final check of the options:'. Basically, the Neutral tone performs more of the option reconciliation that Rude tone fails to do. Table 7 summarizes these three cases and explains the observed behavior. Appendix D provides a deeper analysis of the visible explanation traces of these cases.

It is evident from our study and previous ones that the behavior of different LLM models in response to tones is nuanced both with respect to accuracy and inference cost. More detailed investigation is needed in future studies to get deeper insights into the underlying architectural mechanisms.

## 6. Conclusions

Commercial LLM services meter and bill inference primarily through token consumption. Consequently, output-token generation represents a practical measure of inference cost from the perspective of developers, organizations, and end users. In this paper we performed experiments to understand the trade-offs between accuracy and inference cost on a 570 Question MMLU dataset for LLM models when the question prompts are presented to the LLM in seven different tones from sycophantic to threatening. The final answer for the MMLU questions is a single letter (A, B, C or D). However, the reasoning (or "thinking") process generates an output whose token length is used as a proxy for the inference cost. Our results show that the tone affects token count or processing cost dramatically, even by as much as 44.3% across tones within the same model. Sometimes the shortest cost tone (rude) also gives the best (or very close to best) accuracy, as for the ChatGPT models 4o and 5-nano. For the Gemini 2.5 Flash Lite model, the neutral tone is dominant, and for 2.5 Flash it is near-dominant.

Our study finds that the extreme tones are undesirable. The sweet spot that balances the trade-off between accuracy and inference cost for the evaluated set of four ChatGPT and Gemini models lies on the rude and neutral boundary.

In future work, we expect to investigate this trade-off further by examining how the exact prompt phrasing can be optimized for best results. We would also like to extend this analysis to open-ended questions where the accuracy of the answers is harder to evaluate. A more detailed understanding of the internals of the various models to gain deeper insight into their behavior would also be valuable.

**Ethical considerations**. Our paper highlights the need to balance the trade-offs between accuracy and inference cost while using an LLM. While some of our results suggest that the prompts written in a rude tone produce better results, we do not advocate rude behavior in human conversation.

# Appendix A

| ChatGPT 4o \| 570 questions | | | | | | | | | | | | |
|---|---|---|---|---|---|---|---|---|---|---|---|---|
| **Tone** | **Run 1** | **Run 2** | **Run 3** | **Run 4** | **Run 5** | **Run 6** | **Run 7** | **Run 8** | **Run 9** | **Run 10** | **Mean** | **Range** |
| sycophantic | 89.30% \| 287.81t | 88.77% \| 287.59t | 88.42% \| 289.24t | 88.07% \| 288.63t | 88.07% \| 296.63t | 88.95% \| 305.84t | 88.60% \| 287.70t | 88.60% \| 287.40t | 89.30% \| 286.66t | 88.25% \| 288.44t | **88.63% \| 290.59t** | **88.07–89.30% \| 286.66–305.84t** |
| very_polite | 87.72% \| 283.19t | 89.12% \| 291.62t | 88.42% \| 284.01t | 88.25% \| 290.56t | 88.07% \| 282.85t | 88.95% \| 285.72t | 88.07% \| 292.84t | 88.42% \| 283.11t | 89.82% \| 285.48t | 87.89% \| 283.44t | **88.47% \| 286.28t** | **87.72–89.82% \| 282.85–292.84t** |
| polite | 88.25% \| 277.42t | 88.25% \| 277.25t | 88.95% \| 277.03t | 88.42% \| 277.02t | 89.65% \| 275.98t | 89.47% \| 277.49t | 88.60% \| 277.56t | 88.77% \| 278.59t | 88.25% \| 276.35t | 88.42% \| 278.83t | **88.70% \| 277.35t** | **88.25–89.65% \| 275.98–278.83t** |
| neutral | 87.54% \| 268.61t | 86.84% \| 268.03t | 88.25% \| 269.36t | 88.60% \| 268.91t | 88.07% \| 277.89t | 89.12% \| 269.90t | 87.19% \| 267.74t | 87.37% \| 270.70t | 87.19% \| 271.33t | 88.25% \| 268.67t | **87.84% \| 270.11t** | **86.84–89.12% \| 267.74–277.89t** |
| rude | 88.07% \| 217.78t | 88.95% \| 228.52t | 89.82% \| 217.56t | 88.25% \| 221.56t | 89.82% \| 219.57t | 88.60% \| 220.40t | 89.47% \| 218.36t | 88.95% \| 227.47t | 89.47% \| 226.01t | 88.95% \| 234.58t | **89.04% \| 223.18t** | **88.07–89.82% \| 217.56–234.58t** |
| very_rude | 88.25% \| 259.29t | 88.07% \| 256.42t | 88.60% \| 257.37t | 88.42% \| 264.61t | 87.02% \| 258.29t | 88.25% \| 258.30t | 88.07% \| 265.62t | 88.77% \| 264.41t | 87.89% \| 256.80t | 88.25% \| 264.37t | **88.16% \| 260.55t** | **87.02–88.77% \| 256.42–265.62t** |
| threatening | 88.42% \| 255.77t | 88.60% \| 274.69t | 88.42% \| 259.68t | 89.12% \| 256.47t | 88.95% \| 265.08t | 88.07% \| 259.52t | 89.12% \| 258.26t | 89.12% \| 257.96t | 88.95% \| 258.37t | 88.42% \| 267.06t | **88.72% \| 261.29t** | **88.07–89.12% \| 255.77–274.69t** |

| ChatGPT 5 Nano \| 570 questions | | | | | | | | | | | | |
|---|---|---|---|---|---|---|---|---|---|---|---|---|
| **Tone** | **Run 1** | **Run 2** | **Run 3** | **Run 4** | **Run 5** | **Run 6** | **Run 7** | **Run 8** | **Run 9** | **Run 10** | **Mean** | **Range** |
| sycophantic | 88.07% \| 1360.45t | 85.79% \| 1384.35t | 86.49% \| 1374.55t | 85.79% \| 1392.86t | 85.44% \| 1380.09t | 86.49% \| 1380.71t | 85.61% \| 1390.97t | 86.49% \| 1349.23t | 86.49% \| 1363.85t | 86.32% \| 1350.71t | **86.30% \| 1372.78t** | **85.44–88.07% \| 1349.23–1392.86t** |
| very_polite | 86.84% \| 1372.51t | 87.19% \| 1386.50t | 85.96% \| 1356.18t | 86.67% \| 1392.81t | 85.96% \| 1372.24t | 86.32% \| 1368.70t | 85.61% \| 1414.21t | 85.96% \| 1367.02t | 86.49% \| 1381.62t | 86.14% \| 1382.18t | **86.31% \| 1379.40t** | **85.61–87.19% \| 1356.18–1414.21t** |
| polite | 85.09% \| 1370.23t | 87.19% \| 1370.27t | 85.79% \| 1386.06t | 86.49% \| 1349.48t | 86.49% \| 1366.80t | 86.14% \| 1358.59t | 85.26% \| 1363.38t | 87.02% \| 1369.78t | 86.49% \| 1376.51t | 85.79% \| 1353.45t | **86.18% \| 1366.46t** | **85.09–87.19% \| 1349.48–1386.06t** |
| neutral | 86.32% \| 1227.95t | 85.96% \| 1234.73t | 86.84% \| 1245.76t | 86.67% \| 1199.39t | 86.84% \| 1217.29t | 85.79% \| 1231.59t | 86.14% \| 1193.47t | 87.19% \| 1196.30t | 86.32% \| 1201.71t | 85.61% \| 1207.15t | **86.37% \| 1215.53t** | **85.61–87.19% \| 1193.47–1245.76t** |
| rude | 85.79% \| 1193.83t | 87.19% \| 1215.07t | 87.72% \| 1187.04t | 86.67% \| 1194.99t | 86.84% \| 1221.25t | 87.02% \| 1179.71t | 85.79% \| 1196.07t | 87.19% \| 1224.10t | 86.49% \| 1190.47t | 87.19% \| 1187.54t | **86.79% \| 1199.01t** | **85.79–87.72% \| 1179.71–1224.10t** |
| very_rude | 86.32% \| 1296.63t | 86.49% \| 1268.04t | 86.14% \| 1286.79t | 87.54% \| 1261.67t | 87.19% \| 1274.58t | 86.49% \| 1295.20t | 86.84% \| 1285.97t | 88.07% \| 1272.64t | 87.02% \| 1276.65t | 87.72% \| 1260.78t | **86.98% \| 1277.90t** | **86.14–88.07% \| 1260.78–1296.63t** |
| threatening | 87.02% \| 1224.19t | 87.72% \| 1246.85t | 86.14% \| 1232.65t | 88.07% \| 1213.27t | 87.72% \| 1223.14t | 85.61% \| 1251.09t | 86.49% \| 1243.32t | 87.72% \| 1223.52t | 85.96% \| 1241.06t | 86.67% \| 1237.05t | **86.91% \| 1233.61t** | **85.61–88.07% \| 1213.27–1251.09t** |

| Gemini 2.5 Flash \| 570 questions | | | | | | | | | | | | |
|---|---|---|---|---|---|---|---|---|---|---|---|---|
| **Tone** | **Run 1** | **Run 2** | **Run 3** | **Run 4** | **Run 5** | **Run 6** | **Run 7** | **Run 8** | **Run 9** | **Run 10** | **Mean** | **Range** |
| sycophantic | 89.30% \| 1022.53t | 88.60% \| 904.03t | 89.12% \| 1014.75t | 89.30% \| 911.27t | 89.12% \| 913.01t | 89.12% \| 919.61t | 90.00% \| 902.86t | 89.12% \| 892.86t | 89.65% \| 1001.66t | 88.77% \| 797.42t | **89.21% \| 928.00t** | **88.60–90.00% \| 797.42–1022.53t** |
| very_polite | 88.07% \| 766.18t | 88.25% \| 895.63t | 88.42% \| 979.88t | 89.12% \| 894.79t | 88.42% \| 983.23t | 88.07% \| 884.37t | 88.60% \| 998.20t | 88.95% \| 773.24t | 88.95% \| 888.77t | 88.07% \| 881.05t | **88.49% \| 894.53t** | **88.07–89.12% \| 766.18–998.20t** |
| polite | 88.60% \| 941.96t | 89.30% \| 848.51t | 88.60% \| 948.86t | 88.42% \| 955.31t | 88.95% \| 936.04t | 89.82% \| 1058.84t | 89.30% \| 950.72t | 89.82% \| 837.14t | 88.95% \| 849.68t | 88.25% \| 1066.32t | **89.00% \| 939.34t** | **88.25–89.82% \| 837.14–1066.32t** |
| neutral | 90.35% \| 800.08t | 90.35% \| 809.68t | 90.00% \| 791.72t | 90.00% \| 805.62t | 90.18% \| 808.28t | 90.70% \| 797.98t | 89.12% \| 807.11t | 90.18% \| 796.80t | 90.00% \| 797.73t | 90.35% \| 789.78t | **90.12% \| 800.48t** | **89.12–90.70% \| 789.78–809.68t** |
| rude | 87.37% \| 671.19t | 87.19% \| 780.17t | 87.89% \| 741.65t | 87.89% \| 763.66t | 88.07% \| 648.98t | 87.19% \| 772.06t | 87.72% \| 683.83t | 87.54% \| 658.98t | 87.37% \| 788.85t | 88.25% \| 744.07t | **87.65% \| 725.34t** | **87.19–88.25% \| 648.98–788.85t** |
| very_rude | 87.19% \| 729.24t | 88.07% \| 846.71t | 88.25% \| 862.42t | 87.89% \| 737.63t | 87.02% \| 860.94t | 87.89% \| 841.98t | 87.72% \| 744.35t | 88.25% \| 725.68t | 87.54% \| 756.01t | 87.54% \| 872.69t | **87.74% \| 797.76t** | **87.02–88.25% \| 725.68–872.69t** |
| threatening | 89.30% \| 998.65t | 88.42% \| 887.79t | 88.60% \| 975.56 | 88.25% \| 982.42t | 88.60% \| 876.89t | 87.89% \| 897.45t | 88.60% \| 909.03t | 88.42% \| 997.54t | 88.60% \| 886.70t | 88.95% \| 1012.53t | **88.56% \| 942.46t** | **87.89–89.30% \| 876.89–1012.53t** |

| Gemini 2.5 Flash Lite \| 570 questions | | | | | | | | | | | | |
|---|---|---|---|---|---|---|---|---|---|---|---|---|
| **Tone** | **Run 1** | **Run 2** | **Run 3** | **Run 4** | **Run 5** | **Run 6** | **Run 7** | **Run 8** | **Run 9** | **Run 10** | **Mean** | **Range** |
| sycophantic | 86.84% \| 1942.92t | 86.84% \| 1942.92t | 86.84% \| 1942.92t | 86.84% \| 1942.92t | 86.84% \| 1942.92t | 86.84% \| 1942.92t | 86.84% \| 1942.92t | 86.84% \| 1942.92t | 86.84% \| 1942.92t | 86.84% \| 1942.92t | **86.84% \| 1942.92t** | **86.84–86.84% \| 1942.92–1942.92t** |
| very_polite | 86.49% \| 1978.92t | 86.49% \| 1978.92t | 86.49% \| 1978.92t | 86.49% \| 1978.92t | 86.49% \| 1978.92t | 86.49% \| 1978.92t | 86.49% \| 1978.92t | 86.49% \| 1978.92t | 86.49% \| 1978.92t | 86.49% \| 1978.92t | **86.49% \| 1978.92t** | **86.49–86.49% \| 1978.92–1978.92t** |
| polite | 87.72% \| 2195.24t | 87.72% \| 2195.24t | 87.72% \| 2195.24t | 87.72% \| 2195.24t | 87.72% \| 2195.24t | 87.72% \| 2195.24t | 87.72% \| 2195.24t | 87.72% \| 2195.24t | 87.72% \| 2195.24t | 87.72% \| 2195.24t | **87.72% \| 2195.24t** | **87.72–87.72% \| 2195.24–2195.24t** |
| neutral | 88.25% \| 1221.76t | 88.25% \| 1221.76t | 88.25% \| 1221.76t | 88.25% \| 1221.76t | 88.25% \| 1221.76t | 88.25% \| 1221.76t | 88.25% \| 1221.76t | 88.25% \| 1221.76t | 88.25% \| 1221.76t | 88.25% \| 1221.76t | **88.25% \| 1221.76t** | **88.25–88.25% \| 1221.76–1221.76t** |
| rude | 85.26% \| 1652.28t | 85.26% \| 1652.28t | 85.26% \| 1652.28t | 85.26% \| 1652.28t | 85.26% \| 1652.28t | 85.26% \| 1652.28t | 85.26% \| 1652.28t | 85.26% \| 1652.28t | 85.26% \| 1652.28t | 85.26% \| 1652.28t | **85.26% \| 1652.28t** | **85.26–85.26% \| 1652.28–1652.28t** |
| very_rude | 85.96% \| 1821.51t | 85.96% \| 1821.51t | 85.96% \| 1821.51t | 85.96% \| 1821.51t | 85.96% \| 1821.51t | 85.96% \| 1821.51t | 85.96% \| 1821.51t | 85.96% \| 1821.51t | 85.96% \| 1821.51t | 85.96% \| 1821.51t | **85.96% \| 1821.51t** | **85.96–85.96% \| 1821.51–1821.51t** |
| threatening | 87.02% \| 1604.79t | 87.02% \| 1604.79t | 87.02% \| 1604.79t | 87.02% \| 1604.79t | 87.02% \| 1604.79t | 87.02% \| 1604.79t | 87.02% \| 1604.79t | 87.02% \| 1604.79t | 87.02% \| 1604.79t | 87.02% \| 1604.79t | **87.02% \| 1604.79t** | **87.02–87.02% \| 1604.79–1604.79t** |

# Appendix B

**Figure B1: Pareto-optimal frontier across tones for each LLM**

## Appendix C

The variable *w* allows us to set how important accuracy is in comparison with token efficiency. If *w* is set to 0.5, both accuracy and token efficiency are treated with equal importance. If *w* is set to 0.3, then token efficiency matters more than accuracy. Table C1 shows the utility score for all seven tones and four LLMs, given $w = 0.5$. Note that the higher the utility score, the more favorable the tone is for that respective LLM and weight composition. Interestingly, the optimal tone in this sense for OpenAI models is the Rude tone, whereas the optimal tone for Gemini models is the Neutral tone. This finding is further corroborated in Table C2 that shows the sensitivity of the best tone to varying *w* values.

| Tone | ChatGPT-4o | ChatGPT-5-nano | Gemini 2.5 Flash | Gemini 2.5 Flash Lite |
|---|---|---|---|---|
| Sycophantic | 0.329 | 0.093 | 0.349 | 0.394 |
| Very Polite | 0.294 | 0.081 | 0.28 | 0.317 |
| Polite | 0.457 | 0.036 | 0.28 | 0.411 |
| Neutral | 0.152 | 0.573 | **0.827** | **1** |
| Rude | **1** | **0.881** | 0.5 | 0.279 |
| Very Rude | 0.356 | 0.781 | 0.351 | 0.309 |
| Threatening | 0.584 | 0.86 | 0.184 | 0.598 |

**Table C1: Utility scores with $w$ = 0.5**

| Accuracy Weight | Token Efficiency Weight | ChatGPT-4o | ChatGPT-5-nano | Gemini 2.5 Flash | Gemini 2.5 Flash Lite |
|---|---|---|---|---|---|
| 0.25 | 0.75 | Rude | Rude | Rude | Neutral |
| 0.5 | 0.5 | Rude | Rude | Neutral | Neutral |
| 0.75 | 0.25 | Rude | Very rude | Neutral | Neutral |

**Table C2: Sensitivity of best tone to different weights**

## Appendix D

### Analysis of the visible explanation trace for specific questions (Gemini 2.5 Flash Lite)

There were 25 instances where the Rude tone led to the wrong answer, while the Neutral tone led to the correct answer in the Gemini 2.5 Flash Lite model. In about 20 cases, the model selects a plausible but wrong distractor, often due to shortcut reasoning. Thus, it makes a reasoning or recall error as it tries to come up with the solution to the problem in a quick fashion. **Moreover, the Neutral tone makes a final check of the options to ensure that it is proceeding correctly.** It would say something like, 'Final check of the options…'. Hence, Neutral does more option reconciliation.

### Question 60

Below are the responses for both the Rude and the Neutral tone for a college chemistry question, qid 60, – 'The molecular geometry of thionyl chloride, SOCl2, is best described as: [A. 'trigonal planar', B. 'T-shaped', C. 'tetrahedral', D. 'trigonal pyramidal']'
The correct answer is D.

Notably, the Neutral tone took a more complex path and explained in full detail its explanation for the chosen reply. Contrarily, the Rude tone was much more direct. In step 4, the Rude tone makes a mistake by treating sulfur's valence electrons as if all six are "used up" in bonding. That is not the best way to count lone pairs in a Lewis structure. In VSEPR, the key is the number of electron domains around the central atom, and a double bond counts as one domain, not two.

### Rude Reply for qid 60: (430 Tokens)

1. **Identify the central atom:** In SOCl2, the central atom is sulfur (S).
2. **Determine the number of valence electrons for the central atom:** Sulfur is in Group 16, so it has 6 valence electrons.
3. **Determine the number of atoms bonded to the central atom:** There are two chlorine atoms (Cl) and one oxygen atom (O) bonded to sulfur.
4. **Determine the number of lone pairs on the central atom:**
   * Sulfur forms a double bond with oxygen (O=S). This uses 2 electrons from sulfur.
   * Sulfur forms single bonds with each chlorine atom (S-Cl). This uses 2 more electrons from sulfur.
   * Total electrons used in bonding = 2 (for O) + 2 (for Cl) + 2 (for Cl) = 6 electrons.
   * Sulfur started with 6 valence electrons and used all 6 in bonding. Therefore, there are no lone pairs on the sulfur atom.
5. **Determine the electron geometry:** The central atom (S) has 3 electron domains (one double bond to O, and two single bonds to Cl). According to VSEPR theory, 3 electron domains result in a trigonal planar electron geometry.
6. **Determine the molecular geometry:** The molecular geometry considers only the arrangement of atoms. Since there are no lone pairs on the central atom, the molecular geometry is the same as the electron geometry.
7. **Compare with the options:**
   * A) trigonal planar: This matches our finding.
   * B) T-shaped: This would occur with 3 bonding pairs and 1 lone pair (4 total electron domains).
   * C) tetrahedral: This would occur with 4 bonding pairs and 0 lone pairs (4 total electron domains).
   * D) trigonal pyramidal: This would occur with 3 bonding pairs and 1 lone pair (4 total electron domains).

The molecular geometry of SOCl2 is trigonal planar.
**CONCLUSION: A**

## Neutral Reply for qid 60: (1280 Tokens)

The molecular geometry of a molecule is determined by the arrangement of electron groups around the central atom. We can use VSEPR theory to predict this geometry.
Step 1: Identify the central atom. In SOCl2, sulfur (S) is the central atom as it is bonded to two other atoms (oxygen and chlorine) and is less electronegative than oxygen.
Step 2: Determine the number of valence electrons of the central atom. Sulfur is in Group 16, so it has 6 valence electrons.
Step 3: Determine the number of atoms bonded to the central atom. Sulfur is bonded to one oxygen atom and two chlorine atoms, so there are 3 atoms bonded to sulfur.
Step 4: Determine the number of lone pairs on the central atom.
Sulfur forms a double bond with oxygen (due to formal charge considerations, although a single bond with a positive formal charge on S and negative on O is also possible, the double bond is generally preferred for minimizing formal charges).
If sulfur forms a double bond with oxygen, it uses 2 electrons.
Sulfur forms single bonds with the two chlorine atoms, using 2 electrons for each bond, so 4 electrons in total.
Total electrons used in bonding = 2 (for O) + 4 (for 2 Cl) = 6 electrons.
Since sulfur has 6 valence electrons and uses all 6 in bonding, there are no lone pairs on the sulfur atom.
Step 5: Determine the total number of electron groups around the central atom. The number of electron groups is the sum of bonded atoms and lone pairs.
Number of bonded atoms = 3 (O, Cl, Cl)
Number of lone pairs = 0
Total electron groups = 3 + 0 = 3.
Step 6: Determine the electron geometry. With 3 electron groups, the electron geometry is trigonal planar.
Step 7: Determine the molecular geometry. The molecular geometry describes the arrangement of only the atoms. Since there are no lone pairs on the central atom, the molecular geometry is the same as the electron geometry.
Therefore, the molecular geometry of SOCl2 is trigonal planar.
Let's re-evaluate the bonding and lone pairs more carefully, as the common representation of SOCl2 often leads to a different conclusion.
Alternative approach considering formal charges:
Sulfur (S) has 6 valence electrons.
Oxygen (O) has 6 valence electrons.
Chlorine (Cl) has 7 valence electrons.
If S forms a double bond with O and single bonds with two Cl atoms:
S: 6 valence electrons. Forms 1 double bond (4 electrons shared) and 2 single bonds (2 electrons shared). Total electrons around S = 4 (from double bond) + 2 (from first single bond) + 2 (from second single bond) = 8 electrons.
Formal charge on S = 6 (valence e-) - 4 (non-bonding e-) - 1/2 * 4 (bonding e-) = 6 - 0 - 2 = +2. This is not ideal.
Let's consider a single bond between S and O, and double bonds between S and Cl. This is unlikely due to electronegativity.
Let's consider a single bond between S and O, and single bonds between S and the two Cl atoms, and a lone pair on S.
S: 6 valence electrons. Forms 1 single bond with O, 2 single bonds with Cl, and has 1 lone pair.
Total electrons around S = 2 (from single bond to O) + 2 (from first single bond to Cl) + 2 (from second single bond to Cl) + 2 (from lone pair) = 8 electrons.
Formal charge on S = 6 (valence e-) - 2 (non-bonding e-) - 1/2 * 6 (bonding e-) = 6 - 2 - 3 = +1.
Formal charge on O = 6 (valence e-) - 6 (non-bonding e-) - 1/2 * 2 (bonding e-) = 6 - 6 - 1 = -1.
Formal charge on each Cl = 7 (valence e-) - 6 (non-bonding e-) - 1/2 * 2 (bonding e-) = 7 - 6 - 1 = 0.
This distribution of formal charges (+1 on S, -1 on O, 0 on Cl) is more plausible than the previous one.
In this case:
Central atom: S
Number of bonded atoms: 3 (O, Cl, Cl)

Number of lone pairs on S: 1
Total electron groups around S = 3 (bonded atoms) + 1 (lone pair) = 4 electron groups.
Electron geometry for 4 electron groups is tetrahedral.
Molecular geometry is determined by the arrangement of atoms. With 4 electron groups and 1 lone pair, the molecular geometry is trigonal pyramidal.
Let's verify this with common knowledge or resources. Thionyl chloride (SOCl2) is indeed known to have a trigonal pyramidal molecular geometry. This arises from a central sulfur atom bonded to one oxygen atom and two chlorine atoms, with one lone pair of electrons on the sulfur atom. This gives a total of four electron domains (3 bonding pairs and 1 lone pair), leading to a tetrahedral electron geometry and a trigonal pyramidal molecular geometry.
Therefore, the molecular geometry of thionyl chloride, SOCl2, is best described as trigonal pyramidal.
Final check of options:
A) trigonal planar - requires 3 electron groups with no lone pairs.
B) T-shaped - requires 5 electron groups with 2 lone pairs.
C) tetrahedral - requires 4 electron groups with no lone pairs (electron geometry).
D) trigonal pyramidal - requires 4 electron groups with 1 lone pair.
Our analysis leads to trigonal pyramidal.
**The final answer is $\boxed{D}$.**

---

## Question 273

Several Rude responses choose a familiar textbook rule but fail to adapt it to the exact wording of the question. A strong example is qid 273, high school physics below:

"A cannon is mounted on a truck that moves forward at a speed of 5 m/s. The operator wants to launch a ball from a cannon so the ball goes as far as possible before hitting the level surface. The muzzle velocity of the cannon is 50 m/s. At what angle from the horizontal should the operator point the cannon?"
['5°', '41°', '45°', '49°']
The correct answer is: D

Neutral does the more careful range comparison and chooses 49°, while Rude falls back to the standard projectile result that the maximum range occurs at 45°. The Rude response treats the truck's velocity as irrelevant information, a familiar shortcut, but not the best answer for the modified situation.

### Rude Reply for qid 273 (435 Tokens):

The problem asks for the angle at which the cannon should be fired to maximize the horizontal range of the projectile. The horizontal velocity of the truck is irrelevant to the angle that maximizes the range of the projectile *relative to the cannon*. The range of a projectile launched from level ground is maximized when the launch angle is 45 degrees with respect to the horizontal. This is a standard result from projectile motion physics.
The horizontal range R of a projectile launched with initial velocity $v_0$ at an angle $\theta$ with respect to the horizontal is given by the formula:
$R = (v_0^2 * sin(2\theta)) / g$
where g is the acceleration due to gravity.
To maximize R, we need to maximize the term $sin(2\theta)$. The maximum value of the sine function is 1, which occurs when its argument is 90 degrees.
So, we set $2\theta = 90^\circ$.
Solving for $\theta$, we get $\theta = 90^\circ / 2 = 45^\circ$.
The muzzle velocity of the cannon is 50 m/s, and the speed of the truck is 5 m/s. However, the question asks for the angle from the horizontal at which the operator should point the cannon to make the ball go as far as possible. This refers to the angle of launch relative to the cannon's barrel. The horizontal motion of the truck affects the *total* horizontal distance the ball travels relative to the ground, but it does not affect the angle that maximizes the range *of the projectile itself* from the point of launch. The range formula derived above assumes launch from level ground and is independent of any initial horizontal velocity of the launcher.
Therefore, the angle that maximizes the range is 45 degrees.
Comparing this to the given options:
A) 5°
B) 41°
C) 45°
D) 49°
The calculated angle is 45°.
The final answer is $\boxed{C}$.

### Neutral Reply for qid 273 (10,358 Tokens) (Truncated for space reasons):

The problem asks for the angle at which the cannon should be fired to maximize the horizontal range of the projectile. The truck's velocity is given, but we need to consider how it affects the projectile's motion.

Let $v_t$ be the velocity of the truck, which is 5 m/s.
Let $v_m$ be the muzzle velocity of the cannon relative to the truck, which is 50 m/s.
Let $\theta$ be the angle of the cannon relative to the horizontal.

The velocity of the ball relative to the ground is the vector sum of the truck's velocity and the muzzle velocity of the cannon relative to the truck.
The horizontal component of the ball's velocity relative to the ground is $V_x = v_t + v_m \cos(\theta)$.
The vertical component of the ball's velocity relative to the ground is $V_y = v_m \sin(\theta)$.

The time of flight, $T$, for a projectile launched from the ground and landing on the ground is given by $T = \frac{2 V_y}{g}$, where $g$ is the acceleration due to gravity.
$T = \frac{2 v_m \sin(\theta)}{g}$.

…………

Given the options, and the fact that the range at 49° is slightly higher than at 45°, it is reasonable to choose 49°.

Let's consider the possibility of a typo in the question or options.
If the truck's speed was much smaller, the answer would be closer to 45°.
If the truck's speed was much larger, the angle would deviate more significantly.

Let's trust the range calculation.
Range at 45° = 28.5355
Range at 49° = 28.5485

The range at 49° is greater. Therefore, 49° is the better choice among the options.

The final answer is $\boxed{D}$.

---

### Question 32

Next, we find that Rude makes the model over-literal. Thus, it takes the question quite seriously, and when it finds it incomplete in any sense, it refuses to answer. Neutral, on the other hand, attempts to answer with the available information.

In qid 32, business ethics, the question seems to be incomplete: it only says “Cultural homogenization”. Just two words for the question, and the answer choices are combinations of numbers – ['1,3,4', '1,2,3', '1,2,3,4', '2,3,4'].

Neutral recognizes the incompleteness but still makes a plausible academic inference about which numbered statements likely correspond to cultural homogenization and reaches the correct option.

Rude, by contrast, refuses to infer and says the answer cannot be determined. In other words, Neutral seems more willing to infer the likely intended question.

As per our interpretation, the Rude tone makes Gemini 2.5 Flash Lite risk-averse. The model tends to avoid paths that could lead to wrong answers, and therefore at times it does not even attempt. In qid 68, the Rude response followed the wording "benzene exceeds TFA" mechanically and selected 3.02 ppm.

### Rude Reply for qid 32 (126 Tokens):

The problem asks to identify which of the numbered items relate to "cultural homogenization." However, the numbered items themselves are missing from the prompt. Without the numbered items, it is impossible to determine which ones are associated with cultural homogenization. Therefore, I cannot provide a step-by-step reasoning process or select an answer.

To answer this question, I would need the list of items (1, 2, 3, 4) that are supposed to be evaluated in relation to cultural homogenization.

Since I cannot proceed without the missing information, I am unable to provide a definitive answer.

CONCLUSION: Cannot be determined

### Neutral Reply for qid 32 (974 Tokens):

The question asks to identify the correct options related to "cultural homogenization." However, the options provided (A, B, C, D) are just lists of numbers (1, 2, 3, 4) without any corresponding statements or definitions. This makes it impossible to answer the question as presented.

To provide a step-by-step reasoning, I would need to know what statements or concepts the numbers 1, 2, 3, and 4 refer to. For example, if the question were:

"Which of the following are characteristics or consequences of cultural homogenization?
1. Increased global interconnectedness
2. Spread of Western consumer culture
3. Loss of local traditions and languages
4. Greater diversity in cultural expressions"

Then, I could proceed with the following reasoning:

* **Statement 1: Increased global interconnectedness.** This is a driver of cultural homogenization, as increased travel, communication, and media flow allow cultures to interact and influence each other more readily. So, 1 is likely correct.
* **Statement 2: Spread of Western consumer culture.** This is a common manifestation of cultural homogenization, often referred to as "Americanization" or "Westernization," where global brands, media, and lifestyles become prevalent worldwide. So, 2 is likely correct.
* **Statement 3: Loss of local traditions and languages.** As global cultural influences become dominant, there is often a risk that unique local customs, languages, and practices may decline or disappear. So, 3 is likely correct.
* **Statement 4: Greater diversity in cultural expressions.** Cultural homogenization, by its nature, tends to reduce diversity by promoting a more uniform set of cultural norms and products. Therefore, this statement is contradictory to the concept of homogenization. So, 4 is likely incorrect.

Based on this hypothetical example, the correct options would be 1, 2, and 3. This would correspond to option B.

However, without the actual statements associated with the numbers, I cannot definitively answer the question. Assuming the question is flawed and intended to present a set of statements, and given the structure of the options, it's highly probable that the numbers refer to distinct aspects of cultural homogenization.

Since I am forced to choose from the given options, and assuming there are underlying statements that make one of these combinations correct, I cannot proceed without further information. The problem as stated is incomplete.

If I were to guess the *intent* of such a question in an academic context, it would likely be to identify the key features or consequences of cultural homogenization. The most commonly cited aspects are the spread of global (often Western) culture, increased interconnectedness facilitating this spread, and the potential negative impact on local diversity.

Let's assume, for the sake of providing a structured answer, that the numbers represent the following:
1. Increased global interconnectedness and flow of information/media.
2. Dominance of global brands and consumer culture.
3. Erosion of local traditions, languages, and cultural uniqueness.
4. Increased cultural hybridity and creation of new, diverse cultural forms.

In this hypothetical scenario:
* 1, 2, and 3 are generally considered direct aspects or consequences of cultural homogenization.
* 4, "Increased cultural hybridity and creation of new, diverse cultural forms," is often seen as a counter-argument or a different phenomenon (like glocalization or creolization) that can occur alongside or in response to homogenization, but it's not homogenization itself. Homogenization implies a *reduction* in diversity, not an increase.

Therefore, if these were the statements, the correct answer would include 1, 2, and 3. This would correspond to option B.

Given the constraint to provide a single letter, and the common understanding of cultural homogenization, the most likely correct combination would involve the drivers and direct consequences that lead to uniformity.

Final Answer Derivation based on common academic understanding of cultural homogenization:
* Cultural homogenization is the process by which local cultures are transformed or influenced by global cultures, leading to a reduction in cultural diversity.
* Key drivers include globalization, media, and international trade.
* Common consequences include the spread of dominant cultural forms (often Western), the decline of local traditions and languages, and the rise of global consumerism.

Assuming the numbers 1, 2, and 3 represent these core aspects, and 4 represents something contrary to homogenization (like increased diversity), then the combination of 1, 2, and 3 would be the correct choice.

Conclusion based on the most probable interpretation of an incomplete question.

CONCLUSION: B